# Denoising based on wavelets and deblurring via self-organizing map for Synthetic Aperture Radar images

Mario Mastriani

*Abstract*—This work deals with unsupervised image deblurring. We present a new deblurring procedure on images provided by low-resolution synthetic aperture radar (*SAR*) or simply by multimedia in presence of multiplicative (speckle) or additive noise, respectively. The method we propose is defined as a two-step process. First, we use an original technique for noise reduction in wavelet domain. Then, the learning of a Kohonen self-organizing map (*SOM*) is performed directly on the denoised image to take out it the blur. This technique has been successfully applied to real *SAR* images, and the simulation results are presented to demonstrate the effectiveness of the proposed algorithms.

## I. INTRODUCTION

IN many practical situations, a recorded image presents a noisy and blurred version of an original scene. The image degradation process can be adequately modeled by a linear blur and an additive noise process. Then the degradation model is described by [1]

$$g = D f + n \qquad (1)$$

However, for multiplicative noise, which generally it is called speckle, we propose the follow degradation model

$$g = D f \bullet s \qquad (2)$$

where the vectors $g$, $f$, $n$ and $s$ represent, respectively, the lexicographically (raster scan) ordered noisy blurred image, the original image, the additive noise, and the multiplicative noise (speckle), and the matrix $D$ is the linear degradation process, while the operator "•" means element-by-element multiplication. The image deblurring problem calls for obtaining an estimate of $f$ given $g$ and $D$. For the blind restoration problem, $D$ is not known.

A large number of techniques exist for the denoising [2]-[9] and the deblurring problems [10]-[12]. The image restoration problem is an ill-posed problem. Therefore, a common ingredient in all restoration approaches is that prior information is used in order to restrict the number of possible solutions (basic idea of regularization). Such prior knowledge can be stochastic (i.e., the original image is a sample of a random field) or deterministic (the high frequency energy of the restored image is bounded) in nature. Regularization theory is also applied to the blind restoration problem.

In this paper, an original approach is developed toward both the denoising and the deblurring problems. Such a (non-traditional) approach for denoising is based on the work of Mastriani and Giraldez [13]. They directly apply the Directional Smoothing (*DS*) filter [16] in the Bidimensional Discrete Wavelet Transform (DWT-2D) domain to reduce the presence of speckles, because the edges will be protected from blurring while smoothing. While, in order to face blur generated for the despeckling process, the learning of a Kohonen self-organizing map (*SOM*) is performed directly on the despeckled image. The proposed algorithms differ from the reported results in the literature in a number of ways. Kohonen SOM, for example, is designed from a different point of view than is previously reported in the literature. In the proposed approach, each image to be used for the deblurring problem contains both the low frequency information of the degraded image (the one which is represented by the degraded edges generated for the despeckling process) and the corresponding high frequency information of the original image.

The paper is organized as follows: The Speckle Model is outlined in Section II. The Bayesian Denoising is outlined in Section III. The Directional Smoothing of Coefficients in Wavelet Domain (SmoothShrink) as a denoiser tool is outlined in Section IV. In Section V, we discuss the deblurring method based on Kohonen SOM. In Section VI, the experimental results using the proposed algorithm are presented. Finally, Section VII provides a conclusion of the paper.

## II. SPECKLE MODEL

Speckle noise in SAR images is usually modelled as a purely multiplicative noise process of the form

$I_s(r,c) = I(r,c) S(r,c)$

$= I(r,c) [1+ S'(r,c)]$

$= I(r,c) + N(r,c)$ (3)

The true radiometric values of the image are represented by *I*, and the values measured by the radar instrument are represented by $I_s$. The speckle noise is represented by S. The parameters r and c means row and column of the respective pixel of the image. If $S'(r,c) = S(r,c) - 1$ and $N(r,c) = I(r,c) S'(r,c)$, we begin with a multiplicative speckle $S$ and finish with an additive speckle $N$ [6], which avoid the log-transform, because the mean of log-transformed speckle noise does not equal to zero [15] and thus requires correction to avoid extra distortion in the restored image.

For single-look *SAR* images, *S* is Rayleigh distribu-ted (for amplitude images) or negative exponentially distributed (for intensity images) with a mean of *1*. For multi-look *SAR* images with independent looks, *S* has a gamma distribution with a mean of *1*. Further details on this noise model are given in [16].

### III. BAYESIAN DENOISING

In this section, the denoising of an image corrupted by white Gaussian noise will be considered, i.e.,

$g = x + n$ (4)

where *n* is independent Gaussian noise. We observe *g* (a noisy signal) and wish to estimate the desired signal *x* as accurately as possible according to some criteria. In the wavelet domain, if we use an orthogonal wavelet trans-form, the problem can be formulated as

$y = w + n$ (5)

where *y* noisy wavelet coefficient, *w* true coefficient, and *n* noise, which is independent Gaussian. This is a classical problem in estimation theory. Our aim is to estimate from the noisy observation. The *maximum a posteriori* (*MAP*) estimator will be used for this purpose. Such estimators have been widely advocated for image restoration and reconstruction problems, deriving appro-priately their probability distribution functions (pdf's). Time-variant and time-invariant models will be discussed for this problem in Sections III-A and B, and new MAP estimators are derived.

#### A. Classical Models

The classical *MAP* estimator for (5) is

$\hat{w}(y) = \arg \max_w p_{w|y}(w | y).$ (6)

Using Bayes rule, one gets

$\hat{w}(y) = \arg \max_w [p_{y|w}(y | w) \cdot p_w(w)]$

$= \arg \max_w [p_n(y - w) \cdot p_w(w)].$ (7)

Therefore, these equations allow us to write this estimation in terms of the pdf of the noise ($p_n$) and the pdf of the signal coefficient ($p_w$). From the assumption on the noise, $p_n$ is zero mean Gaussian with variance $\sigma_n$, i.e.,

$p_n(n) = \frac{1}{\sigma_n \sqrt{2\pi}} \cdot \exp\left(-\frac{n^2}{2\sigma_n^2}\right).$ (8)

It has been observed that wavelet coefficients of natural images have highly non-Gaussian statistics [7]-[9]. The pdf for wavelet coefficients is often modeled as a generalized (heavy-tailed) Gaussian [7]-[9].

$p_w(w) = K(s, p) \cdot \exp\left(-\left|\frac{w}{s}\right|^p\right).$ (9)

where *s*, *p* are the parameters for this model, and *K*(*s*, *p*) is the parameter-dependent normalization constant. Other pdf models have also been proposed [7]-[9]. In practice, generally, two problems arise with the Bayesian approach when an accurate but complicated pdf $p_w(w)$ is used: 1) It can be difficult to estimate the parameters of $p_w$ for a specific image, especially from noisy data, and 2) the estimators for these models may not have simple closed form solution and can be difficult to obtain. The solution for these problems usually requires numerical techniques [17]-[19].

Let us continue developing the *MAP* estimator and show it for Gaussian and Laplacian cases. Equation (7) is also equivalent to

$\hat{w}(y) = \arg \max_w [\log(p_n(y - w)) + \log(p_w(w))].$ (10)

As in [7]-[9], let us define $f(w) = \log(p_w(w))$. By using (8), (10) becomes

$\hat{w}(y) = \arg \max_w [-\frac{(y-w)^2}{2\sigma_n^2} + f(w)].$ (11)

This is equivalent to solving the following equation for $\hat{w}$ if $p_w(w)$ is assumed to be strictly convex and differentiable.

$\frac{y - \hat{w}}{\sigma_n^2} + f'(\hat{w}) = 0.$ (12)

If $p_w(w)$ is assumed to be a zero mean gaussian density with variance $\sigma^2$, then $f(w) = -\log(\sqrt{2\pi}\sigma) - w^2/2\sigma^2$, and the estimator can be written as

$$\hat{w}(y) = \frac{\sigma^2}{\sigma^2 + \sigma_n^2} \cdot y. \qquad (13)$$

If it is Laplacian

$$p_w(w) = \frac{1}{\sqrt{2}\sigma} exp\left(-\frac{\sqrt{2}|w|}{\sigma}\right). \qquad (14)$$

then $f(w) = -\log(\sigma\sqrt{2}) - \sqrt{2}|w|/\sigma$, and the estimator will be

$$\hat{w}(y) = sign(y)\left(|y| - \frac{\sqrt{2}\sigma_n^2}{\sigma}\right)_+. \qquad (15)$$

Here, $(g)_+$ is defined as

$$(g)_+ = \begin{cases} 0, & \text{if } g < 0 \\ g, & \text{otherwise} \end{cases} \qquad (16)$$

Equation (15) is the classical soft shrinkage function. Let us define the soft operator as

$$soft(g,\tau) = sign(g) \cdot (|g| - \tau)_+. \qquad (17)$$

The soft shrinkage function (15) can be written as

$$\hat{w}(y) = soft\left(y, \frac{\sqrt{2}\sigma_n^2}{\sigma}\right). \qquad (18)$$

*B. Mask Convolution Model*

If pdf is

$$p_w(w) = exp\left(-\frac{(M^{-1} - I)\hat{w}^T \hat{w}}{2\sigma_n^2}\right). \qquad (19)$$

where $M$ is a matrix that represents a convolution mask and $I$ is the identity matrix, then

$$f(\hat{w}) = -\frac{(M^{-1} - I)\hat{w}^T \hat{w}}{2\sigma_n^2}. \qquad (20)$$

with

$$f'(\hat{w}) = -\frac{(M^{-1} - I)\hat{w}}{\sigma_n^2} \qquad (21)$$

Replacing (21) into (12), the estimator can be written as

$$\hat{w}(y) = M \cdot y. \qquad (22)$$

where Eq.(22) represents a 2D-convolution between the mask $M$, and the noisy wavelet coefficients $y$.

If the image has $R$ rows and $C$ columns, and the kernel (mask) has $r$ rows and $c$ columns, then the size of the output image will have $R$-$r$+1 rows, and $C$-$c$+1 columns.

Mathematically we can write the convolution as:

$$\hat{w}(i,j) = \sum_{p=1}^{r}\sum_{q=1}^{c} y(i + p - 1, j + q - 1) M(p,q) \qquad (23)$$

where $i$ runs from 1 to $R$-$r$+1 and $j$ runs from 1 to $C$-$c$+1.

That is to say, the novel method was deduced for all kind of convolutional mask filter, however, we choose the directional smoothing because protect the edges from blurring while smoothing the wavelet coefficients.

IV. DIRECTIONAL SMOOTHING OF COEFFICIENTS IN WAVELET DOMAIN (SMOOTHSHRINK)

Since Sveinsson *et al.* [20] directly apply the Enhanced-Lee filter in the Bidimensional Discrete Wavelet Transform (DWT-2D) domain to reduce the presence of speckles, then and with the same approach, we use the DS [14], because the edges will be protected from blurring while smoothing. The experimental results demonstrate that DS is better than Enhanced-Lee filter in all the carried out experiments.

Therefore, we begin decomposing the speckled SAR image into four wavelet subbands: Coefficients of Approximation (CA), and speckled coefficients of Diagonal Detail (CDD$_s$), Vertical Detail (CVD$_s$), and Horizontal Detail (CHD$_s$), respectively. We apply DS within each high subband, and reconstruct a SAR image from the modified wavelet coefficients, that is to say, despeckled coefficients of Diagonal Detail (CDD$_d$), Vertical Detail (CVD$_d$), and Horizontal Detail (CHD$_d$), respectively, as shown in Fig. 1, where: IDWT-2D is the inverse of DWT-2D. Based on Eq.(3) SmoothShrink does not need log-transform [6].

*A. Theory of Directional Smoothing*

To protect the edges from blurring while smoothing, a directional averaging filter must be applied. Spatial averages $d(r,c:\Theta)$ are calculated in several directions as shown in the follow equation

$$d(r,c:\Theta) = \frac{1}{N_\Theta} \sum_{k \in W\Theta} \sum_{l \in W\Theta} x(r-k, c-l) \qquad (24)$$

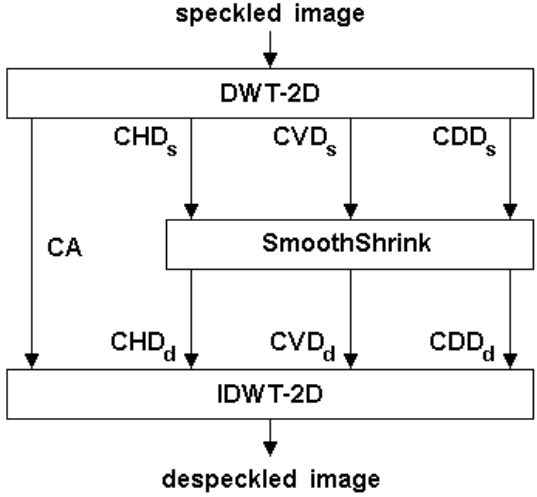

Fig. 1 Smoothing of Coefficients in wavelet domain (SmoothShrink)

and a direction $\Theta^*$ is found such that $|x(r,c) - d(r,c:\Theta^*)|$ is minimum, where $x$ is the respective detail subband. Then

$$d(r,c) = d(r,c:\Theta^*) \qquad (25)$$

gives the desired result for the suitably chosen window $W$, $N_\Theta$ is the number of directions, and $k$ and $l$ depends on the size of such windows (kernel) [21].

The *DS* filter has a speckle reduction approach that performs spatial filtering in a square-moving window defined as kernel, and is based on the statistical relationship between the central pixel and its surrounding pixels as shown in Fig. 2.

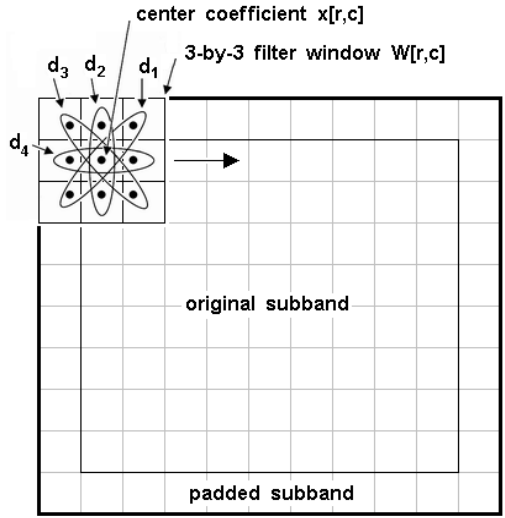

Fig. 2. 3-by-3 filter window on a subband (CHD, CVD, and CDD)

The size of the filter window can range from 3-by-3 to 33-by-33, with an odd number of cells in both directions. A larger filter window means that a larger area of the image will be used for calculation and requires more computation time depending on the complexity of the filter's algorithm. If the size of filter window is too large, the important details will be lost due to over smoothing. On the other hand, if the size of the filter window is too small, speckle reduction may not be very effective. In practice, a 3-by-3 or a 7-by-7 filter window usually yields good results in the cases under study [22].

DS performs the filtering based on either local statistical data given in the filter window to determine the noise variance within the filter window, or estimating the local noise variance using the effective equivalent number of looks (*ENL*) [13], [22] of the image under study. The estimated noise variance is then used to determine the amount of smoothing needed for each subimage. The noise variance obtained from the local filter window is more applicable if the backscatter of an area is constant (flat and homogeneous) whilst *ENL* is suitable if there are difficulties determining if an area of the image is flat.

Most simple nonlinear thresholding rules for wavelet based denoising assume that the wavelet coefficients are independent [17]-[19]. However, wavelet coefficients of natural images have significant dependencies. In this paper, we will consider the dependencies between the coefficients and their neighbors in detail. The Smooth Shrink do not assume the independence of wavelet coefficients, because, It is based on the *DS* algorithm, which keeps in mind the incidence of the neighboring elements by means of the employment of a mask, which can be observed in the algorithm that is detailed next.

*B. Algorithm*

Algorithm represents DS function for four directions and a 3x3 kernel in MATLAB® code

```
function x = ds(x)

[ROW,COL] = size(x);

for r = 2:ROW-1
  for c = 2:COL-1
    d(1) = (x(r,c-1)  +x(r,c)+x(r,c+1)  )/3;
    d(2) = (x(r-1,c)  +x(r,c)+x(r+1,c)  )/3;
    d(3) = (x(r-1,c-1)+x(r,c)+x(r+1,c+1))/3;
    d(4) = (x(r+1,c-1)+x(r,c)+x(r-1,c+1))/3;
    for n = 1:4
      D(n) = abs(d(n)-x(r,c));
    end
    [Dmin,aDmin] = min(D);
    x(r,c) = d(aDmin);
  end
end
```

where:
**x** represents the bitmap matrix of the image
ds(•) is the function that calculate the directional
    smoothing of (•)
size(•) is the function that calculate the dimensions of
    matrix (•)
ROW is the number of rows and COL is the number of
    columns of **x**.
**d** represents the vector of directions

**D** represents the vector of absolute differences
abs(•) is the function that calculate the absolute value of (•)
min(•) is the function that calculate the minimum of vector (•) and its location
Dmin is the minimum of vector **D**
aDmin is the location of Dmin

## V. Deblurring Via Self-Organizing Map

### A. Self-Organizing Map

Researches on neurobiology have shown that centers of diverses activities as thought, speech, vision, hearing, lie in specific areas of the cortex and these areas are ordered to preserve the topological relations between informations while performing a dimensionality reduction of the representation space. Such organization led Kohonen to develop the SOM algorithm [23]. This kind of competitive neural network is composed of one or two dimensional array of processing elements or neurons in the input space. All these neurons receive the same inputs from external world. Learning is accomplished by iterative application of unlabeled input data. As training process, the neurons evolve in the input space in order to approximate the distribution function of the input vectors. After this step, large-dimensional input vectors are, in a sense, projected down on the one or two-dimensional map in a way that maintains the natural order of the input data. This dimensional reduction could allow us to visualize and to use easily, on a one or two-dimensional array, important relationships among the data that might go unnoticed in a high-dimensional space.

The model of *SOM* used in our application is a one-dimensional array of $n$ nodes. To each neuron $N_i$, a weight vector $w_i = (w_{i1}, w_{i2}, ...., w_{ip})^t \in \Re^p$ is associated.

During learning procedure, an input vector $x \in \Re^p$ randomly selected among vectors of the training set, is connected to all neurons in parallel. The input $x$ is compared with all the neurons in the Euclidean distance sense *via* variable scalar weight $w_{ij}$. At the $k$th step, we assign the vector $x$ to the winning or leader neuron $N_l$ if:

$$\left\| x - w_l^{[k]} \right\| = \min_i \left\| x - w_i^{[k]} \right\| \qquad (26)$$

All the neurons within a certain neighborhood around the *leader* participate in the weight-update process. Considering random initial values for $w_i^{[0]}$ ($0 \leq i \leq n$), this learning process can be described by the following iterative procedure:

$$w_i^{[k+1]} = w_i^{[k]} + H_{li}^{[k]} (x^{[k]} - w_i^{[k]}) \qquad (27)$$

The lateral interactions among topographically close elements are modeled by the application of *a neighbourhood function* or a smoothing Kernel defined over the winning neuron [23]. This Kernel can be written in terms of the Gaussian function

$$H_{li}^{[k]} = \alpha^{[k]} exp\left(-\frac{d^2(l,i)}{2(\sigma^{[k]})^2}\right) \qquad (28)$$

where $d(l, i) = \|l - i\|$ is the distance between the node $l$ and $i$ in the array, $\alpha^{[k]}(t)$ is the learning-rate factor and $\sigma^{[k]}$ defines the width of the Kernel at the iteration $k$. For the convergence, it is necessary that $H_{li}^{[k]} \rightarrow 0$ when $k \rightarrow T$, where $T$ is the total number of step of the process [23]. Therefore, for the first step, $\alpha^{[k]}$ should start with a value that is close to unity, thereafter decree-sing monotonically [23]. To achieve this task, we use

$$\alpha^{[k]} = \alpha^{[0]} \left(1 - \frac{k}{T}\right) \qquad (29)$$

Moreover, as learning proceeds, the size of the neighbourhood should be diminished until it encompasses only a single unit. So, we applied for the width of the Kernel the monotonically decreasing function:

$$\sigma^{[k]} = \sigma^{[0]} \left(\frac{\sigma^{[T-1]}}{\sigma^{[0]}}\right)^{k/(T-1)} \qquad (30)$$

The ordering of the map occurs during the first steps, while the remaining steps are only needed for the fine adjustment of the weight values.

### B. Iterative Learning

The learning process is performed directly on the real image to be deblurred. An input vector is filled with the grey levels of the pixels of the image, see Fig. 3. Therefore, each neuron has *rows-by-columns* weights allowing to locate it in the input space. At each step, the weights are modified according to Eq. (27). Experiments have shown that this training strategy provides as good results as an ordered image scanning process while spending less processing time.

$\sigma$ has a significant impact on the quality of the convergence. We have to start with a fairly large value to globally order the map. The initial value $\sigma_0$ of $\sigma$ p can be half the length of the network. During learning, $\sigma$ has to decrease monotonically until it reaches a small value. Experiments have shown that $\sigma_{T-1} = 0.1$ is a good choice and provides the minimum quantization error defined by

$$E_{quant} = \left\| x - w_l \right\| \qquad (31)$$

where $w_l$ is the weight vector associated to the leader neuron of the input vector $x$ after learning step.

*C. SOM Deblurring*

The deblurring task consists in using the Eq.(27) over the image. For each iteration, the corresponding input vector $x$ is compared with all the neurons using Eq. (26). The *winning neuron*, the one which leads to the smallest distance, gives the class of the winner pixel in which iteration. However, before any deblurring task, we have to *calibrate* the map in order to associate the label *mean* or *edges* to each neuron.

Assuming that the input vector $x_0 = (0, ...., 0)^t$ should represent an image setting on a identical mean value, it is very useful to define the distance graph representing the Euclidean distance in the rows-by-columns-dimensional space between the point $x_0$ and all the neurons. Such a graph is given in Fig. 4 and Fig. 5 respectively before and after learning for a 512-by-512-neuron network.

Both these figures show that the maximal distance between two successive cells is widely smaller after learning than before. We show only 100 of 512x512 neurons around the winner. We can deduce that, after learning, neurons that are topologically close in the array are close in the input space too. As a matter of fact, neurons that are physical neighbors should respond to a similar input vectors [10].

Experiments have shown that *Equant* is a monotonically decreasing function of the number of steps and reaches an asymptotic value for large value of *T*. Similar to [10], one hundred times the number of network units seems to be a reasonable compromise solution between speed and quality of learning.

VI. EXPERIMENTAL RESULTS

Several experiments have been performed in studying the effectiveness of the proposed algorithms. In this section, we first describe some of the experimental results obtained with the proposed image restoration algorithm in Section IV, and then some of the experimental results obtained with the proposed blur identification algorithm in Section V.

*A. Experiment 1*

In this case, we dealt with an image degraded by the Gaussian degradation function and additive zero-mean Gaussian noise, as shown in Fig.6. The Gaussian degradation function is expressed as follows:

$$h(i, j) = K \, exp\left(-\frac{i^2 + j^2}{2\sigma^2}\right) \qquad (32)$$

where *K* is a normalizing constant ensuring that the blur is of unit volume and $\sigma^2$ is the variance. We experimented with values of $\sigma^2$ equal to 1.5 and 3.5 and two levels of noise resulting in values of blurred *SNR* (*BSNR*) of 20 and 10 dB. The BSNR in dB for an *I* x *J* image is defined by [11]

$$BSNR = 10\,log_{10}\left\{\frac{\frac{1}{IJ}\sum_{i,j}[D\,f - E\{g\}]^2}{\sigma_n^2}\right\} \qquad (33)$$

where *D* and *f* are respectively the blurring function and the original image, shown in (3), $E\{g\}$ is the expected value of the degraded image and $\sigma_n^2$ the variance of the noise. For comparison, we present deblurring results obtained by a VQ-based blind image restoration algorithm [11], and Constrained Least Squares (*CLS*) filter. The last one is a widely used linear restoration technique [1], [24]-[26].

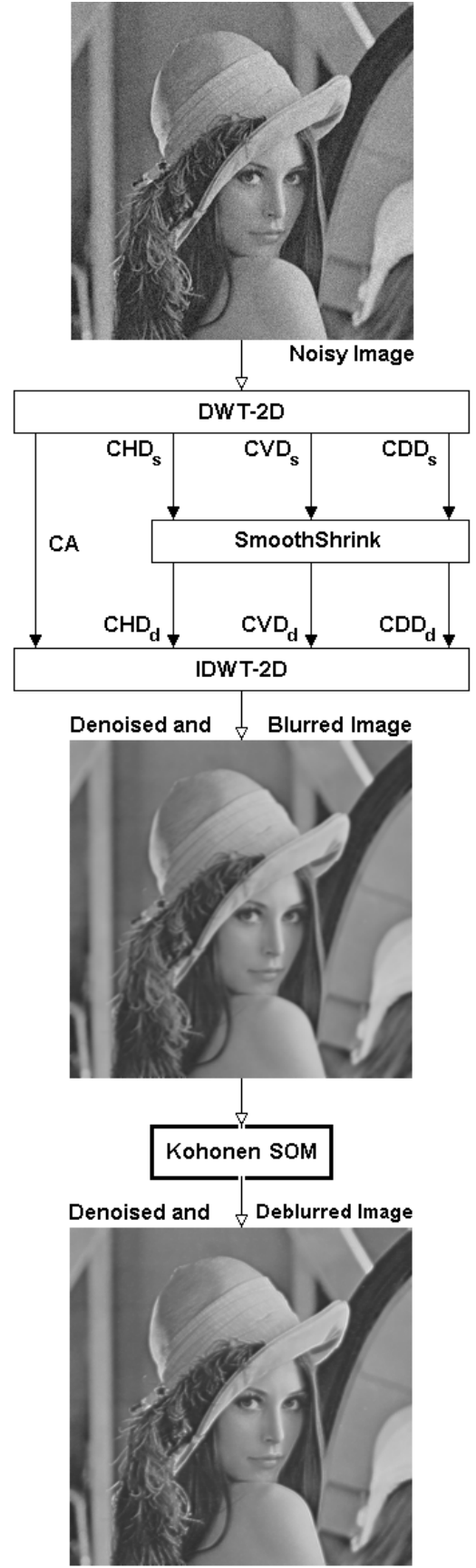

Fig. 3 Denoising and deblurring.

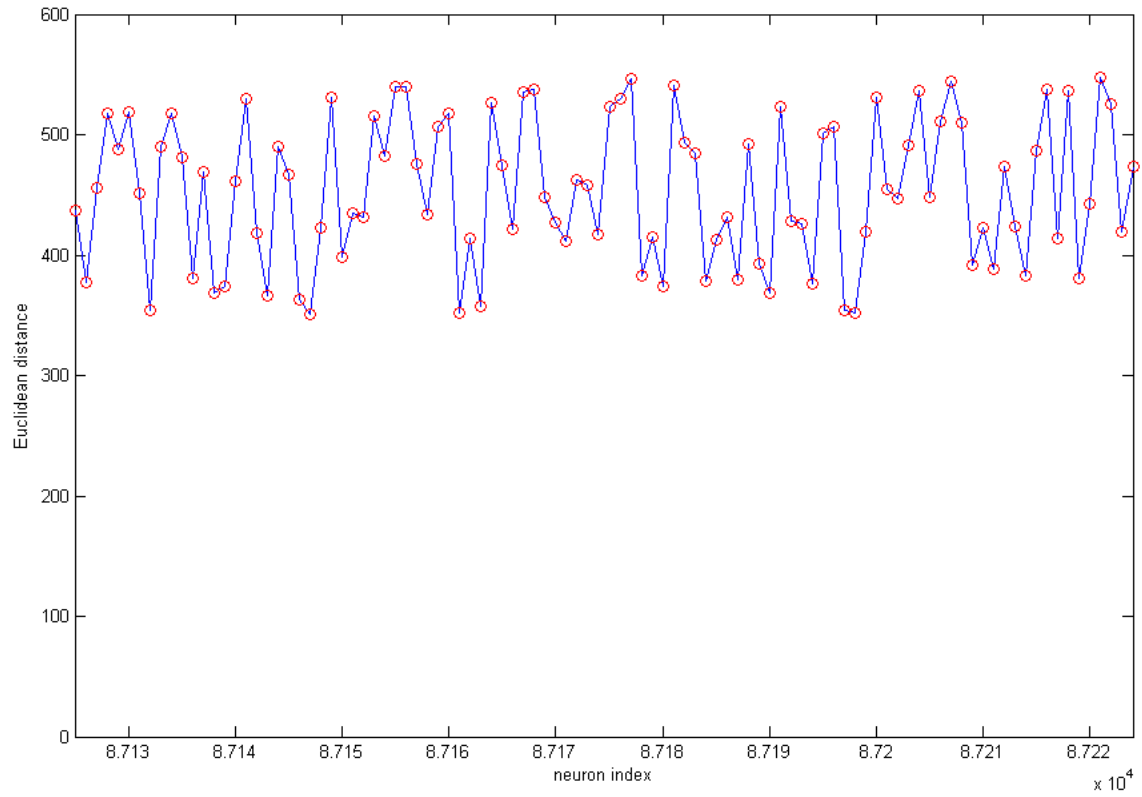

Fig. 4 The distance graph between the 100 neurons of the SOM before learning. One dimensional curve of neurons.

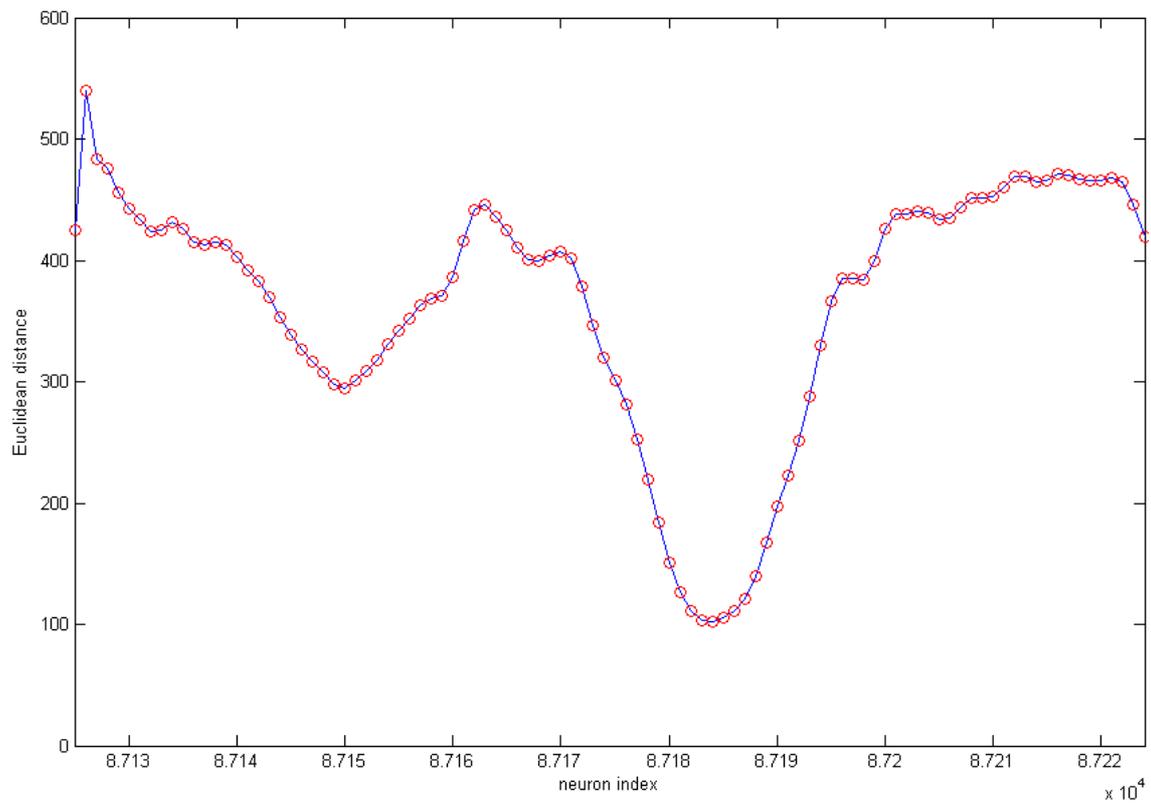

Fig. 5 The distance graph between neurons obtained after learning. One dimensional curve of neurons.

A 3 x 3 Laplacian filter is used to implement the CLS filter and the regularization parameter is chosen as the reciprocal of the *BSNR* of the observed image. The choice of *1/BSNR* can be justified through the set-theoretic formulation of the problem, where the loose bounds on the constraints we try to satisfy are the noise variance and the high frequency signal variance [27].

Only one type of image is used for the experimental results reported here. It is an 8 bit/pixel, 512 x 512 pixels, natural image (Lena), which is often used for the evaluation of image-processing algorithm. As a quantitative metric of the performance of the restoration algorithm, we used the improvement in *SNR* (*ISNR*) defined by

$$ISNR = 10\,log_{10}\left\{\frac{\sum_{i,j}|f(i,j) - g(i,j)|^2}{\sum_{i,j}|f(i,j) - \hat{f}(i,j)|^2}\right\} \qquad (32)$$

where $f(i,j)$, $\hat{f}(i,j)$ and $g(i,j)$ are the original, restored and degraded images, respectively. Although the ISNR metric reflects the global properties of the restoration and may not fully reflect the subjective improvement of the image quality, it is useful for providing an objective means to measure and compare the quality of the results.

Finally, the comparative study is shown in Table I, where, RP means regularization parameter [11].

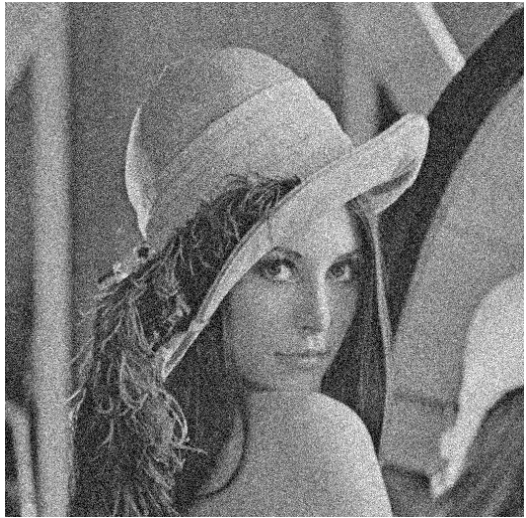

(a)

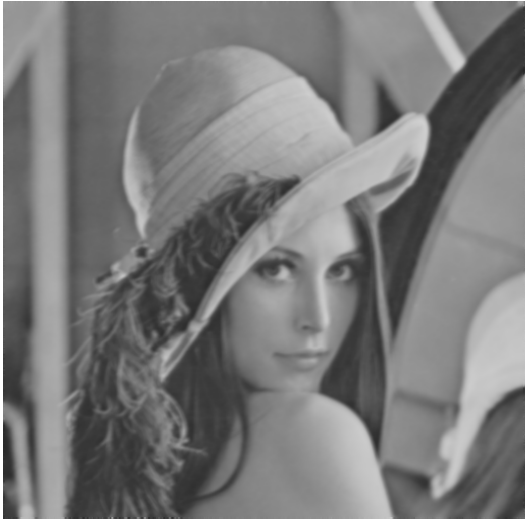

(b)

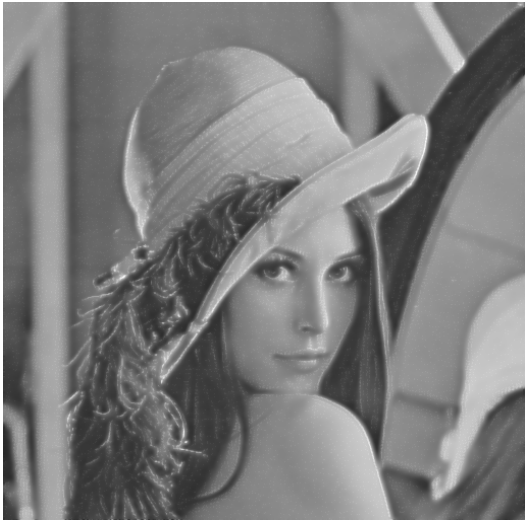

(c)

Fig. 6 a) Noisy image, b) blurred image by denoised filtering, and c) deblurred image by SOM.

TABLE I
COMPARATIVE STUDY BASED ON ISNR

| Metric | Restored Image by | | | |
|---|---|---|---|---|
| | SOM | VQ | CLS RP = 0.05 | CLS RP = 0.1 |
| ISNR | 3.08 dB | 2.91 dB | 2.47 dB | 0.91 dB |

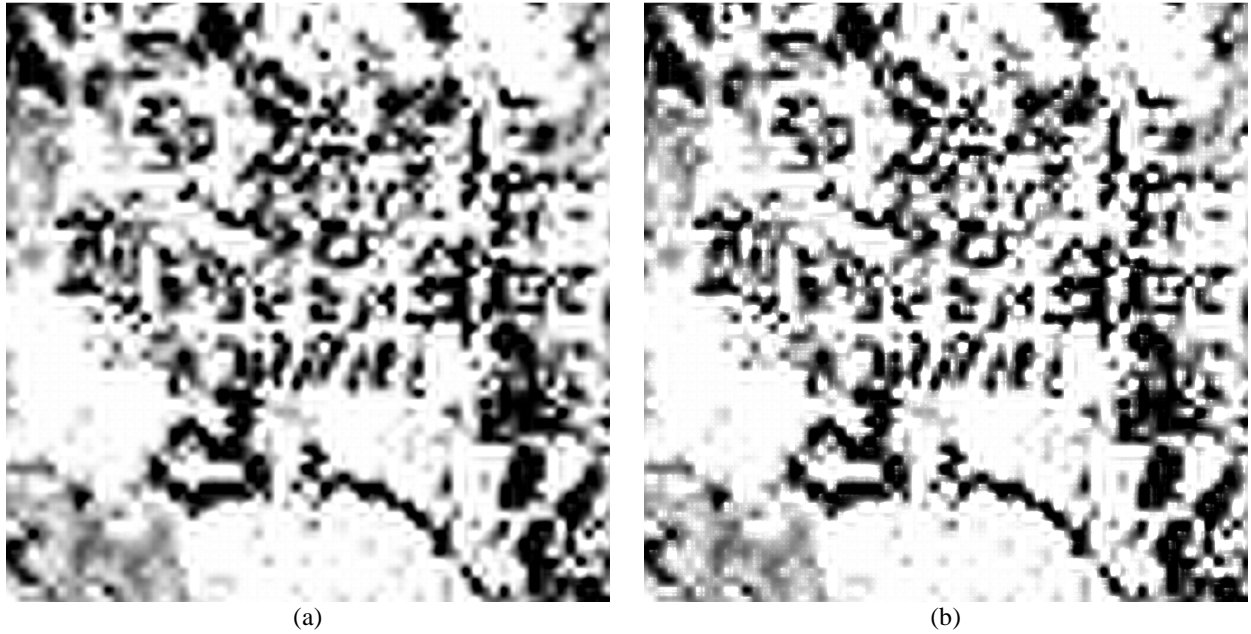

Fig. 7 a) Original low-resolution image from satellite ERS-2 without noise and, b) enhanced resolution of image by SOM.

## B. Experiment 2

In this case, we dealt with an original low-resolution image from satellite ERS-2, see Figure 7. The combined action of SmoothShrink [13] and SOM allows it to obtain an enhanced resolution of image, with an improvement in the ISNR, as shown in Table II.

TABLE II
COMPARATIVE STUDY BASED ON ISNR

| Metric | Restored Image by | | | |
|---|---|---|---|---|
| | SOM | VQ | CLS RP = 0.05 | CLS RP = 0.1 |
| ISNR | 2.15 dB | 1.72 dB | 1.45 dB | 1.17 dB |

## C. Employed Algorithm

SOM in MATLAB® code

```
function I = som(I)

[ROW,COL] = size(I);
sigma = input('sigma = ');
alfa = input('alfa = ');
d = floor(sigma/2);

for r = 1+d:ROW-d
  for c = 1+d:COL-d
    Sigma = I(r-(1+d)+1:r-(1+d)+sigma,
              c-(1+d)+1:c-(1+d)+sigma);
    e = I(r,c)-mean2(Sigma);
    I(r,c) = I(r,c)+alfa*e;
  end
end
```

where:
*I* represents the bitmap matrix of the image
som(•) is the function that calculate the *SOM* of (•)
size(•) is the function that calculate the dimensions of matrix (•)
*ROW* is the number of rows and *COL* is the number of columns of *I*.
*sigma* = $\sigma$, and defines the width of the Kernel
*alfa* = $\alpha$, is the learning-rate factor

## VII. CONCLUSIONS

Although the SOM-based deblurrer
1) did not use the traditional Gaussian neighbourhood function as a property for the algorithm,
2) the learning-rate factor is constant along iterations, and
3) the width of the Kernel is constant along iterations too,

the results are better than the results of such well-known methods as deblurring.

An input vector if filled with the grey levels of the pixels contained in a $\sigma \times \sigma$ pixels window sliding over the image. Therefore, each neuron has $\sigma \times \sigma$ weights allowing to locate it in the input space. Contrary to the Yao *et al*'s method, at each step, the location of the window in the image is determined in an ordered sliding form. Experiments have shown that this training strategy provide a better results with less processing time and a more simply code.

The main drawback of applying the combination of wavelets and SOM on low-resolution synthetic aperture radar (SAR) images is the sum of their respective computational complexities.

Finally, the natural extension of this work is in medical applications, as well as in microarrays denoising.